\documentclass{article}

\usepackage{arxiv}

\usepackage[utf8]{inputenc} 
\usepackage[T1]{fontenc}    
\usepackage[colorlinks=true,allcolors=black]{hyperref}       
\usepackage{url}            
\usepackage{booktabs}       
\usepackage{amsfonts}       
\usepackage{nicefrac}       
\usepackage{microtype}      
\usepackage{lipsum}		
\usepackage{graphicx}
\usepackage[numbers]{natbib}
\usepackage{doi}
\usepackage{pifont}
\usepackage{xcolor}
\usepackage{amsmath}
\usepackage{fancyhdr}
\usepackage{array}
\usepackage{multirow}
\usepackage{colortbl}
\usepackage{hyperref}
\usepackage{booktabs}
\usepackage{array}
\usepackage{amssymb}
\usepackage{enumitem} 
\usepackage{amsmath}
\usepackage{subcaption}

\title{From Text to Visuals: Using LLMs to Generate Math Diagrams with Vector Graphics}

\author{Jaewook Lee\\
University of Massachusetts Amherst\\
jaewooklee@cs.umass.edu\\
\And
Jeongah Lee\\
University of Massachusetts Amherst\\
jeongahlee@cs.umass.edu \\
\And
Wanyong Feng\\
University of Massachusetts Amherst\\
wanyongefeng@cs.umass.edu\\
\And
Andrew Lan\\
University of Massachusetts Amherst\\
andrewlan@cs.umass.edu
}

\date{}

\renewcommand{\headeright}{}
\renewcommand{\undertitle}{Preprint, under review}

\begin{document}
\maketitle

\begin{abstract}
Advances in large language models (LLMs) offer new possibilities for enhancing math education by automating support for both teachers and students. While prior work has focused on generating math problems and high-quality distractors, the role of visualization in math learning remains under-explored. Diagrams are essential for mathematical thinking and problem-solving, yet manually creating them is time-consuming and requires domain-specific expertise, limiting scalability.
Recent research on using LLMs to generate Scalable Vector Graphics (SVG) presents a promising approach to automating diagram creation. Unlike pixel-based images, SVGs represent geometric figures using XML, allowing seamless scaling and adaptability. Educational platforms such as Khan Academy and IXL already use SVGs to display math problems and hints. In this paper, we explore the use of LLMs to generate math-related diagrams that accompany textual hints via intermediate SVG representations. We address three research questions: (1) how to automatically generate math diagrams in problem-solving hints and evaluate their quality, (2) whether SVG is an effective intermediate representation for math diagrams, and (3) what prompting strategies and formats are required for LLMs to generate accurate SVG-based diagrams.
Our contributions include defining the task of automatically generating SVG-based diagrams for math hints, developing an LLM prompting-based pipeline, and identifying key strategies for improving diagram generation. Additionally, we introduce a Visual Question Answering-based evaluation setup and conduct ablation studies to assess different pipeline variations. By automating the math diagram creation, we aim to provide students and teachers with accurate, conceptually relevant visual aids that enhance problem-solving and learning experiences.
\end{abstract}


\section{Introduction}

Advances in large language models (LLMs) have the potential to significantly advance math education, by automating support for both teachers and students. From the teacher’s perspective, LLMs can not only generate math problems~\cite{mitra2024orca,shridhar2022automatic} but also high-quality distractors~\cite{feng2024exploring,fernandez2024divert,scarlatos2024improving} in multiple-choice questions. 
From the student’s perspective, LLM-based tutors, such as Khan Academy’s Khanmigo~\cite{khanmigo}, can engage learners in dialogues and provide real-time feedback.

Despite this progress, helping students visualize math concepts remains an under-explored, yet potentially impactful form of student support. Visualization plays a powerful role in mathematical thinking and can help students grasp abstract mathematical ideas~\cite{arcavi2003role,mayer2024learning,presmeg2020visualization}. In particular, \emph{diagrams} can play a crucial in math problem-solving. Research shows that encouraging students to draw and interact with diagrams strengthens their mathematical reasoning ability~\cite{uesaka2010effects}. Providing students with opportunities to use diagrams also boosts understanding, builds proficiency, and increases confidence~\cite{uesaka2007kinds}.

In practice, teachers and students often resort to drawing diagrams by hand, which can be time-consuming, prone to errors, and less adaptable to quick changes or personalization. Therefore, visualization is not as scalable to a large number of students as other support mechanisms. Although specialized software systems can render mathematical diagrams, most require domain-specific programming skills~\cite{ye2020penrose}, making them difficult to be widely adopted.

\sloppy
Recent research on using LLMs to generate Scalable Vector Graphics (SVG)~\cite{ferraiolo2000scalable} offers a promising approach to automating diagram generation. Unlike pixel-based images, SVGs represent geometric figures and text using XML elements~\cite{bray2008xml}, enabling it to scale seamlessly across different screen sizes. Educational technology platforms, such as Khan Academy\cite{khanacademy} and IXL~\cite{IXLLearning}, leverage SVGs for displaying math problems and hints. LLMs’ have shown emerging ability to not only understand but also generate SVG code~\cite{cai2023leveraging,nishina2024svgeditbench}. In this paper, we explore how to use this capability for student support with \emph{hints}: we explore using LLMs to automatically generate math-related diagrams that accompany textual hints, via intermediate SVG code representations. By automating the creation of such diagrams, we envision a future where teachers and students can easily access visually accurate, conceptually relevant illustrations that support them during math learning. We raise the following research questions (RQs):
\begin{itemize}
    \item \textbf{RQ1}: How do we automatically generate math diagrams in math hints during problem solving, and how do we evaluate the quality of diagrams?
    \item \textbf{RQ2}: Is SVG an effective intermediate representation of math diagrams?
    \item \textbf{RQ3}: What are the key prompting strategies and format required for LLMs to generate accurate and effective SVG-based diagrams?
\end{itemize}

\subsection{Contributions.} In this paper, we define the task of automatically generating SVG-based diagrams to accompany textual hints to support students during math problem-solving. We explore an LLM prompting-based pipeline and identify the key strategies necessary to improve their performance on this task. We also develop a Visual Question Answering (VQA)~\cite{antol2015vqa}-based setup to evaluate the effectiveness of the generated diagrams. We also conduct several ablation studies to answer our research questions, by exploring variations of the pipeline to assess their impact on diagram quality and accuracy.

\section{Task Definition}
Our task is to generate mathematical diagrams based on textual feedback for math problems. Formally, each problem has a statement \(\mathcal{P}\) and a corresponding set of hints \(\mathcal{H}\) that guides students through step-by-step solutions. The problem statements are mostly textual, but may contain diagrams as well. Each hint step $i$ in $\mathcal{H}$ can be decomposed into a pair ${(T_i, D_i)}$, where $T_i$ represents the textual hint and $D_i$ represents the corresponding diagram in SVG format. The textual hint always exists but the diagram may not; in that case, $D_i = \emptyset$. We note that the diagrams are not just pixel-based images, but rather a textual representation in SVG format, ensuring that diagrams are stored and processed as structured code rather than raster or vector image files.

\begin{figure}[h]
    \centering
    \includegraphics[width=1\textwidth]{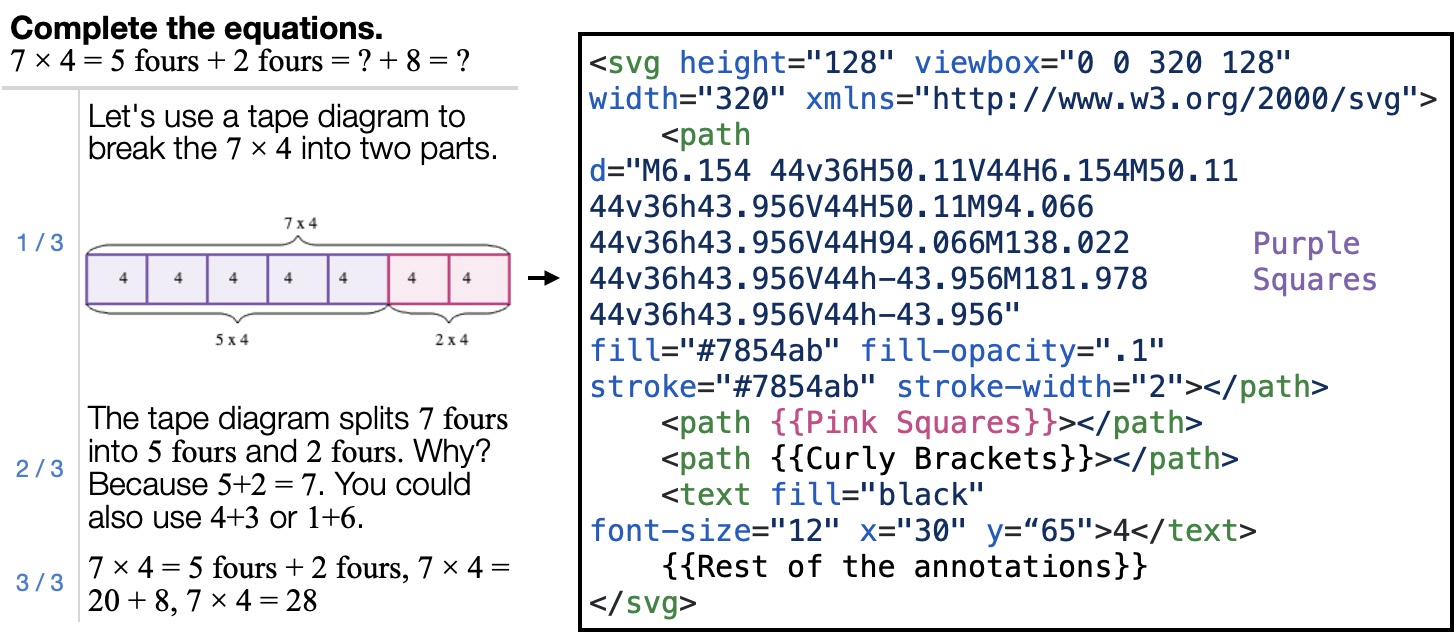} 
    \caption{A math problem on the topic ``distributive property of multiplication''. The left side shows the problem, and step-by-step hints, while the right side shows the diagram in the first-step hint in SVG format.}
    \label{fig:svg_example}
\end{figure}

Figure~\ref{fig:svg_example} shows a math problem on the topic of learning distributive property of multiplication, with a diagram illustrating the decomposition of \(7 \times 4\) into \(5 \times 4\) and \(2 \times 4\). The SVG format relies on XML-based elements, utilizing \texttt{<path>} for shapes and \texttt{<text>} for annotations. Distinct \texttt{<path>} elements highlight the segmented multiplications, with different colors differentiating \(5 \times 4\) and \(2 \times 4\). A third \texttt{<path>} outlines the overall structure, incorporating curved separators for clarity. Additionally, labels and mathematical expressions (\(7 \times 4\), \(5 \times 4\), and \(2 \times 4\)) are included within the blocks.

\textbf{Goal.} Given the problem statement \(\mathcal{P}\) and the previously provided hints \(\mathcal{H}_{<i} = \{(T_0, D_0), (T_1, D_1), \dots, (T_{i-1}, D_{i-1})\}\), the objective is to generate the next hint diagram \(D_i\) corresponding to the provided textual feedback \(T_i\).

\section{Methodology}
\label{sec:method}
\begin{figure}[h]
    \centering
    \includegraphics[width=0.8\textwidth]{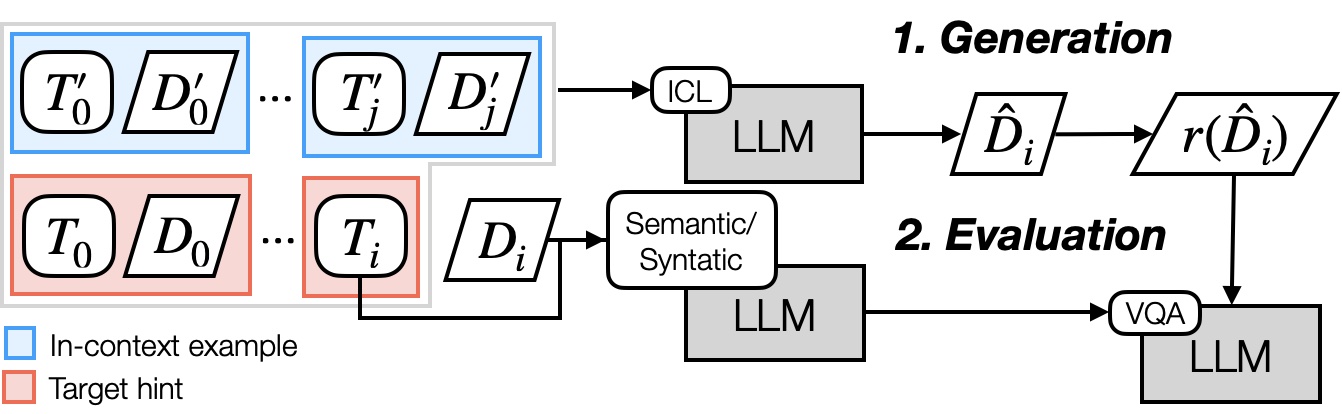} 
    \caption{An overview of our pipeline for vector image generation and evaluation. $r(\cdot)$ represents rasterization, where a vector image is rendered into a pixel image. Problem statement $\mathcal{P}$ is removed due to spatial constraints. Rounded boxes correspond to text $T$, parallelograms correspond to diagrams $D$, either in SVG code format or pixel-based image format.}
    \label{fig:pipeline}
\end{figure}

In this section, we detail our pipeline that generates and evaluates hint diagrams using LLMs, as shown in Figure~\ref{fig:pipeline}. We use LLMs in three ways: one for generating the diagram and two for evaluating the quality of the generated diagram. 

\subsection{Diagram Generation}

For diagram generation, we employ in-context learning (ICL)~\cite{dong2024survey}, selecting examples from the same topic as the target problem. The prompt is structured to include four key components: (1) a task description, (2) instructions for generating diagrams, (3) an ICL example, and (4) the target problem.

The task description specifies that the goal is to generate diagrams in SVG format, ensuring that the diagram aligns with textual feedback. The instructions provide guidelines for generating accurate and meaningful diagrams. The ICL example consists of a problem \(\mathcal{P'}\) from the same topic as the target problem \(\mathcal{P}\), accompanied by hints \(\mathcal{H'} = \{(T'_0, D'_0), \dots, (T'_{j}, D'_{j})\}\). Finally, for the target problem, we follow the same structure but provide only the last textual input \(T_i\), aiming to generate the predicted diagram \(\hat{D_i}\). 

\paragraph{Pipeline Variants}
As a preliminary step, we first evaluate a problem that contains only a single hint diagram. We also study variants of the pipeline that differs from the overall pipeline on three aspects: (1) Diagram format (2) Textual hint generation (3) On-task demonstration. 
For diagram format, we explore automatically generating pixel images using generative text-to-image (T2I) models. Specifically, we extend the prompt to include a description of the ICL example (e.g., \textit{``A diagram of seven rectangles, ...''} for the diagram in Figure~\ref{fig:svg_example}). The LLM then generates a description of the intended SVG code, which serves as a prompt for a T2I model (e.g., DALL·E 3~\cite{openai2023dalle3}). The resulting generated image, along with the description, tests the importance of using SVG code to represent diagrams for LLMs to comprehend.
For textual hint generation, rather than generating only the diagram \(D_i\) given the textual hint \(T_i\), we generate first \(T_i\) and then \(D_i\). This setup tests whether LLMs can understand math hints or can only generates diagrams according to exact specifications.
For on-task demonstration, we study whether providing the diagram from the previous hint step (\(D_{i-1}\)) influences the generation of the current diagram (\(D_i\)) to align with the ground truth. Since diagrams at consecutive hint steps often resemble each other, this setup tests whether the LLM can maintain logical consistency in diagram evolution rather than treating each step as an independent generation task.

\section{Evaluation}
\label{sec:vqa}

Measuring the quality of diagrams in SVG format is challenging, since there can be multiple ways to represent the same image with different SVG code. Moreover, code-level similarity evaluation can be misleading since SVGs are scalable and can be moved or rotated without altering their appearance. Even small changes in position, size, or structure can make one diagram look different from another in code, even if they appear identical. Furthermore, while direct visual comparison with a ground truth diagram may seem intuitive, it often fails to assess whether a diagram effectively conveys the intended information.

\begin{table}[ht!]
\centering
\caption{Semantic and syntactic evaluation criteria we use to generate visual questions to evaluate the quality of math hint diagrams.}
\renewcommand{\arraystretch}{1} 
\setlength{\tabcolsep}{6pt} 
\begin{tabular}{@{}p{0.45\textwidth} p{0.45\textwidth}@{}}
\toprule
\textbf{Semantic Evaluation} & \textbf{Syntactic Evaluation} \\
\toprule
\begin{itemize}[leftmargin=*, noitemsep, topsep=0pt]
  \item Clarity of core concepts
  \item Facilitation of understanding
  \item Alignment with math reasoning
  \item Ease of interpretation
  \item Minimization of ambiguity
\end{itemize}
&
\begin{itemize}[leftmargin=*, noitemsep, topsep=0pt]
  \item Inclusion of key components
  \item Accurate numerical and labels
  \item Structural consistency
  \item Completeness 
  \item Adherence to math notations
\end{itemize}
\\
\bottomrule
\end{tabular}
\label{tab:evaluation_criteria}
\end{table}


To address these limitations, we employ a VQA approach using LLMs to analyze the quality of generated diagrams. Given their proven ability to solve geometry problems in pixel-based images~\cite{gao2023g,zhang2024mavis}, LLMs serve as an effective tool for this analysis. We generate semantic and syntactic questions based on the evaluation criteria in Table~\ref{tab:evaluation_criteria}, using the ground truth diagram \(D_i\). We distinguish between \textbf{semantic} (meaning and relationships) and \textbf{syntactic} (structural arrangement) aspects, capturing both the meaning the diagram conveys and how it is visually composed. 

For example, for the diagram in Figure~\ref{fig:svg_example}, the semantic evaluation checks whether the diagram correctly conveys the distributive property by illustrating how \(7 \times 4\) can be broken down into \(5 \times 4\) and \(2 \times 4\), as suggested by the hint text (e.g., \textit{``...break the \(7 \times 4\) into two parts.''}). This evaluation also assesses whether the diagram provides clear visual cues, such as distinct labeling of each group, to help learners understand the decomposition. In contrast, the syntactic evaluation examines whether the tape diagrams are structured in a way that accurately represents this breakdown. It verifies that the \(5 \times 4\) and \(2 \times 4\) segments are proportional and well-labeled, that partitions align correctly with the intended grouping, and that no structural inconsistencies (e.g., misaligned segments or missing markers) compromise the clarity of the illustration.  

To the best of our knowledge, no prior VQA task has evaluated the quality of math diagrams. Therefore, we heuristically define the evaluation criteria in Table~\ref{tab:evaluation_criteria}. These evaluation questions, generated based on the criteria and ground truth \(D_i\), along with the VQA task and the rasterized version of the generated diagram \(r(\hat{D_i})\), form the evaluation prompt, as shown in Figure~\ref{fig:pipeline}. The LLM for the VQA task evaluates each question independently, providing a Yes or No answer along with reasoning for its response. We later validate this VQA-based approach by comparing results for both ground truth and generated diagrams, along with the LLM's reasoning, to validate the VQA process. 

\section{Experimental Evaluation}
In this section, we detail the experiment conducted using our pipeline and discuss various aspects of using LLMs to generate hint diagrams. For all experiments, we utilize GPT-4o~\cite{openai2024gpt4o} (temperature: 0.5) for both generation and evaluation. Notably, the model's multimodal capabilities are utilized exclusively for evaluation, specifically in the VQA task for assessing the generated diagrams.
%

\begin{table}[]
    \centering
    \caption{Five math topics we use in the experiment, along with corresponding problems, hints, and diagrams.}
    \renewcommand{\arraystretch}{1.2} 
    \setlength{\tabcolsep}{6pt}      
    \scalebox{1}{%
    \begin{tabular}{m{2cm} m{2.5cm} m{4cm} m{2cm}}
        \toprule
        \textbf{Topic} & \textbf{Problem} & \textbf{Textual Hint} & \textbf{Diagram} \\
        \midrule
        Divide by 2 
            & $6 \div 2 = \square$ 
            & If we split 6 circles into 2 equal rows, how many circles are in each row? 
            & \multirow{2}{*}{\includegraphics[width=0.15\textwidth]{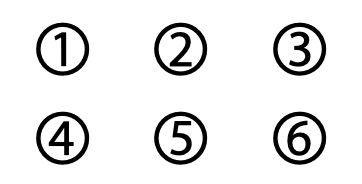}} \\\cmidrule(lr){1-3}

        Multiple by 2 (or 4)
            & $2 \times 3 = \square$ 
            & We can think of $2 \times 3$ as $2$ rows of $3$ circles. How many circles are there? 
            & \\

        \midrule
        Multiply by 1 (or 0) 
            & $1 \times 3 = \square$ 
            & We can think of $1 \times 3$ as $1$ group of $3$ circles: How many circles are there? 
            & \includegraphics[width=0.15\textwidth]{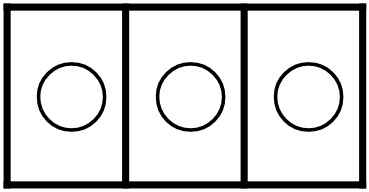} \\

        \midrule
        Comparing fractions
            & Compare using $>$, $<$, or $=$:  $\frac{1}{4} \square \frac{2}{4}$ 
            & Which fraction shows the most shaded area? 
            & \includegraphics[width=0.15\textwidth]{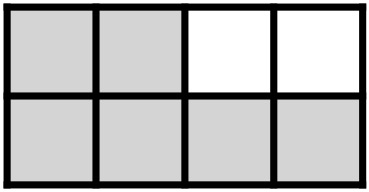} \\
        \midrule
        Area Formulation Intuition 
            & Area of $X$? 
            & There are 2 rows of 2 square units each. 
            & \includegraphics[width=0.15\textwidth]{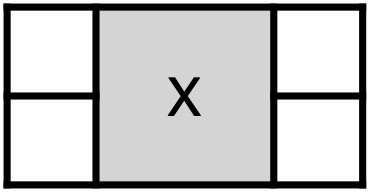} \\
        \bottomrule
    \end{tabular}
    }
    \label{tab:cat}
\end{table}

\subsection{Experimental Design}
\label{sec:experimental_design}

We start by manually reviewing diagrams from Khan Academy for grades 1-3. Since these grades introduce simpler mathematical concepts, they are an ideal starting point to explore using LLMs to generate math problem diagrams. 
The most common diagram type (among number lines, bar graphs, etc.) is the \textit{array structure}. These diagrams contain objects that are arranged in rows and columns to represent multiplication, grouping, or spatial relationships. They play a key role in early math education, helping students understand concepts such as repeated addition, equal grouping, and area models. 
%


Within array-structured diagrams, we focus on cases where only one hint in the sequence includes a diagram, as stated above in Section~\ref{sec:method}; Table~\ref{tab:cat} shows several examples. Focusing on a single-hint diagram enables us to assess how well the pipeline generates visual representations without relying on previous hint diagrams. By starting with this simpler setting, we first evaluate LLM capabilities before expanding to more complex variations. Additionally, we conduct an ablation study with multiple diagrams to explore how additional visuals impact comprehension and the pipeline’s overall effectiveness.


Our selected problems cover math topics such as division and multiplication by a factor, fraction comparisons, and area formulation, each with at least six problems per topic. We use one of them as the ICL example, and evaluate our pipeline using the others. This setup captures diversity in both problem complexity within each topic and different object types within the array structure, including numbered circles, grouped circles, colored rectangles, and grouped rectangular areas. This diversity enhances the robustness of our evaluation. 

\section{Results and Discussion}
In this section, we detail experimental results and discuss our findings.

\subsection{RQ1: Overall Performance}

Table~\ref{tab:vector_gt_comparison} shows the performance of our pipeline across different math topics (top half) and for different variants of our pipeline (bottom half). We show the percentage of VQs answered correctly for both the ground truth diagram and generated diagrams. 
We select five problems from each topic and compute the average VQA accuracy. As detailed above in Section~\ref{sec:vqa}, the generated questions are based on the ground truth vector images. Therefore, in theory, the ground truth diagram should always receive a score of 1. We calculate accuracy by dividing the VQA accuracy of the generated diagram by that of the ground truth, for both semantic and syntactic evaluation. 
We see that, as expected, the ground truth outperforms the predicted diagrams in both semantic (\(0.85\) vs. \(0.80\)) and syntactic (\(0.79\) vs. \(0.76\)) evaluations. 
%
%
Overall, these results demonstrate that the pipeline generates diagrams comparable to the ground truth in the VQA task. The pipeline, which uses LLMs with ICL prompting and SVG code as an intermediate representation, appears to be a viable approach for diagram generation.

\begin{figure}[h]
    \centering
    \begin{subfigure}{0.45\textwidth}
        \centering
        \includegraphics[width=\linewidth]{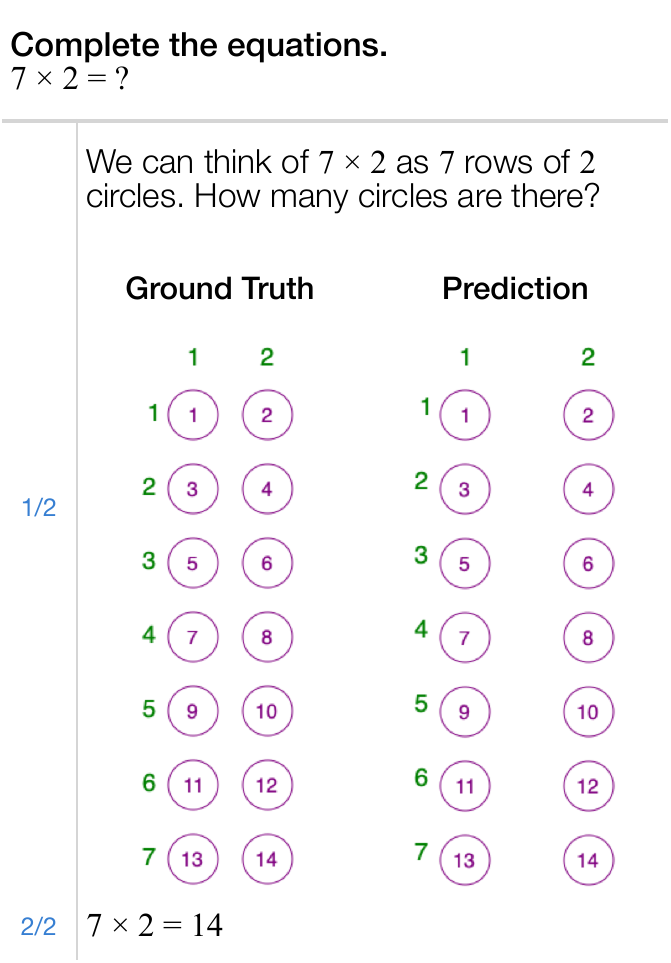}
        \caption{Multiply by 2 (or 4)}
        \label{fig:good_case}
    \end{subfigure}
    \hfill
    \begin{subfigure}{0.45\textwidth}
        \centering
        \includegraphics[width=\linewidth]{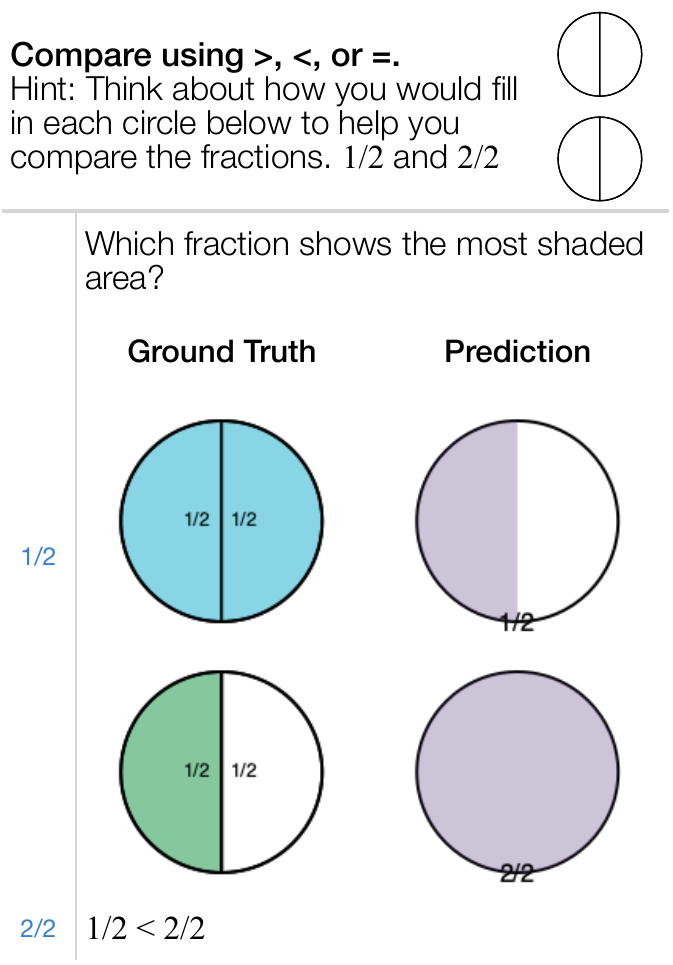}
        \caption{Comparing Fractions}
        \label{fig:bad_case}
    \end{subfigure}
    \caption{Two math problems and their corresponding generated diagrams in a hint step, showing cases of successful and unsuccessful diagram generation.}
    \label{fig:case_studies}
\end{figure}

However, there are some discrepancies;
Figure~\ref{fig:case_studies} shows two examples, one where the ground truth scores perfectly (1.0) and the other where it performs relatively poor. See Table~\ref{tab:evaluation-scoring} for the VQA questions alongside their results on both images. We discuss them in more detail below. 

\begin{table}[h]
    \centering
    \caption{VQA evaluation results for diagrams generated by our pipeline and a breakdown over topics (top part). Ablation studies with pipeline variants to test the importance of its components (bottom part), with scores averaged across topics. An asterisk (*) indicates cases where the problem involves multiple diagrams and we evaluate only the last diagram.}
    \setlength{\tabcolsep}{5pt}
    \begin{tabular}{lcccccc}
        \toprule
        \multirow{2}{*}{\textbf{Topic}} & \multicolumn{2}{c}{\textbf{Ground Truth}} & \multicolumn{2}{c}{\textbf{Prediction}} & \multicolumn{2}{c}{\textbf{Accuracy}} \\
        \cmidrule(lr){2-5} \cmidrule(lr){6-7}
        & Sem. & Syn. & Sem. & Syn. & Sem. & Syn. \\
        \midrule
        Divide by 2 & 1.00 & 1.00 & 1.00 & 0.88 & - & -  \\
        Multiple by 2 (or 4) & 1.00 & 0.96 & 0.92 & 0.84 & - & - \\
        Multiply by 1 (or 0) & 1.00 & 1.00 & 0.96 & 0.92 &  - & -  \\
        Comparing Fractions & 0.72 & 0.48 & 0.56 & 0.56 &  - & -  \\
        Area Formulation Intuition & 0.52 & 0.52 & 0.56 & 0.60 &  - & -  \\
        \midrule
        \midrule
        Pipeline & 0.85 & 0.79 & 0.80 & 0.76 & 0.94 & 0.96\\
        Pipeline (T2I) &  & & 0.12 & 0.10 & 0.14 & 0.13 \\
        Pipeline (Textual Hint) &  &  & 0.74 & 0.68 & 0.87 & 0.86 \\
        Pipeline* (On-Task)  & 0.93 & 0.81 & 0.96 & 0.79 & 1.03 & 0.97 \\
        \bottomrule
    \end{tabular}
    \label{tab:vector_gt_comparison}
\end{table}


\begin{table}[h]
    \centering
    \caption{Semantic and syntactic questions for the topic in Figure~\ref{fig:case_studies}, where T denotes the ground truth diagram and G denotes the generated diagram.}
\scalebox{.8}{
    \begin{tabular}{p{1cm} p{6cm} p{0.5cm} p{0.5cm} p{6cm} p{0.5cm} p{0.5cm}}
        \toprule
        & \multicolumn{3}{c}{\textbf{Multiply by 2 (or 4)}} 
        & \multicolumn{3}{c}{\textbf{Comparing Fractions}} \\
        \cmidrule(lr){2-4} \cmidrule(lr){5-7}
        & \textbf{Question} & \textbf{T} & \textbf{G}
        & \textbf{Question} & \textbf{T} & \textbf{G} \\
        \midrule
        
        \multirow{5}{*}{\parbox{1.5cm}{ \textbf{Sem.}}}
        & Does the diagram clearly show 7 rows of 2 circles each? & \ding{51} & \ding{51} 
        & Does the diagram clearly represent the fractions 1/2 and 2/2? & \ding{55} &  \ding{51}\\

        & Is the arrangement of circles in the diagram easy to interpret? & \ding{51} & \ding{51}
        & Does the diagram effectively use shading to differentiate the fractions? & \ding{55} &  \ding{51}\\

        & Does the diagram effectively illustrate the concept of multiplication as repeated addition? & \ding{51} & \ding{51}
        & Is the diagram aligned with the hint text by showing the shaded areas for comparison? &  \ding{51}& \ding{51} \\

        & Are the circles in the diagram labeled in a way that supports understanding the total count? & \ding{51} & \ding{51}
        & Can students easily interpret the shaded and unshaded portions in the diagram? &  \ding{55}&  \ding{51}\\

        & Does the diagram align with the text feedback by visually representing 7 rows of 2 circles? & \ding{51}  & \ding{51}
        & Does the diagram avoid any misleading or ambiguous elements in representing the fractions? & \ding{51} &  \ding{51}\\
        
        \midrule
        
        \multirow{5}{*}{\parbox{1.5cm}{ \textbf{Syn.}}}

        & Does the diagram accurately show 7 rows of circles as mentioned in the text feedback? & \ding{51}  & \ding{51}
        & Does the diagram include both circles as described in the hint text? &  \ding{55} & \ding{51} \\
        
        & Are there exactly 2 circles in each row as described in the text feedback? & \ding{51} & \ding{51} 
        & Are the fractions labeled correctly as 1/2 and 2/2 in the diagram? &  \ding{55} & \ding{51} \\
        
        & Are the circles in the diagram correctly labeled from 1 to 14? & \ding{51} & \ding{51}
        & Is the shading in the diagram accurately representing the fractions as described? & \ding{55} & \ding{51} \\
        
        & Is the alignment and positioning of the circles consistent with the description in the text feedback? & \ding{51} & \ding{51}
        & Are the circles in the diagram properly aligned and positioned according to the standard layout? & \ding{55} & \ding{51} \\
        
        & Does the diagram follow standard mathematical notation and conventions for representing multiplication? & \ding{51} & \ding{51}
        & Does the diagram follow standard mathematical notation in representing the fractions and shaded areas? & \ding{55} & \ding{51}\\

        \bottomrule
    \end{tabular}
}
    \label{tab:evaluation-scoring}
\end{table}

\subsubsection{Case Studies}
We now use two examples to illustrate when our pipeline excels and fails. Figure~\ref{fig:good_case} shows a positive example, where the textual hint encourages students to interpret \(7 \times 2\) as seven rows of two circles each, reinforcing the concept of multiplication as repeated addition. The ground truth diagram visually represents the idea with a \(7 \times 2\) grid of circles, numbered to indicate a total of 14 circles. The diagram generated by our pipeline closely mirrors the structure, preserving all essential components and colors from the ground truth, with the only difference being slight spacing variations between the first and second columns. The high degree of similarity is further validated by the VQA results, which are entirely correct. These results also show that VQA effectively evaluates vector-based diagrams rasterized in pixel images, capturing their conceptual equivalence in a way that direct SVG-level code comparisons does not.

Figure~\ref{fig:bad_case} shows a negative example, where a diagram is provided as a hint to help students compare \(\frac{1}{2}\) and \(\frac{2}{2}\). The textual hint encourages students to color the given area to visually represent each fraction. The ground truth diagram depicts the two fractions in different colors. Additionally, each half-circle is labeled as \(\frac{1}{2}\) to reinforce the concept that \(\frac{2}{2} = \frac{1}{2} + \frac{1}{2}\), demonstrating that \(\frac{2}{2}\) is greater than \(\frac{1}{2}\).  However, the diagram generated by our pipeline primarily follows the textual information provided in the problem statement, without breaking down \(\frac{2}{2}\) into \(\frac{1}{2} + \frac{1}{2}\). The VQA results also reflect this observation, as the VQs evaluates on the presence of labels on \(\frac{2}{2}\), leading to all syntactic questions being incorrect for the ground truth diagram.

These examples highlight strengths and limitations of our pipeline. While the generated diagrams generally align well with expected representations, as seen in the multiplication case, discrepancies emerge in nuanced problems like fraction comparisons. These differences underscore the importance of ensuring diagrams fully convey intended pedagogical objectives; future work should focus on refining diagram generation to better align with these educational goals.

\subsection{Pipeline Variants}

\subsubsection{RQ2: Using a T2I Model} 

We conduct a first ablation experiment and explore generating pixel-based
diagrams directly, using a T2I model (DALL·E 3), bypassing the SVG code as an intermediate step. We found that in most cases, the generated diagrams are not useful, which is reflected in low VQA accuracy numbers in Table~\ref{tab:vector_gt_comparison}. We show one example in Figure~\ref{fig:good_case}. While LLMs effectively generate textual descriptions for SVG diagrams, the T2I model generates low-quality outputs not suitable for hint diagrams. Most generated diagrams contain arbitrary numbers within the description and exhibit unintended three-dimensional characteristics, whereas all ground truth diagrams are strictly two-dimensional. This observation is not surprising, since prior research shows that diffusion models struggle with generating high-quality geometric images~\cite{huang2024autogeo}. 
\begin{table}[htbp]
    \centering
    \caption{Generated pixel-based diagram for the problem in Figure~\ref{fig:good_case}, along with the textual description used to generate the diagram using a T2I model.}
    \renewcommand{\arraystretch}{1.1} 
    \begin{tabular}{m{0.16\textwidth} | p{0.75\textwidth}} 
        \hline
        \textbf{Pixel Image} & \textbf{Description} \\
        \hline
        \includegraphics[width=0.1\textwidth]{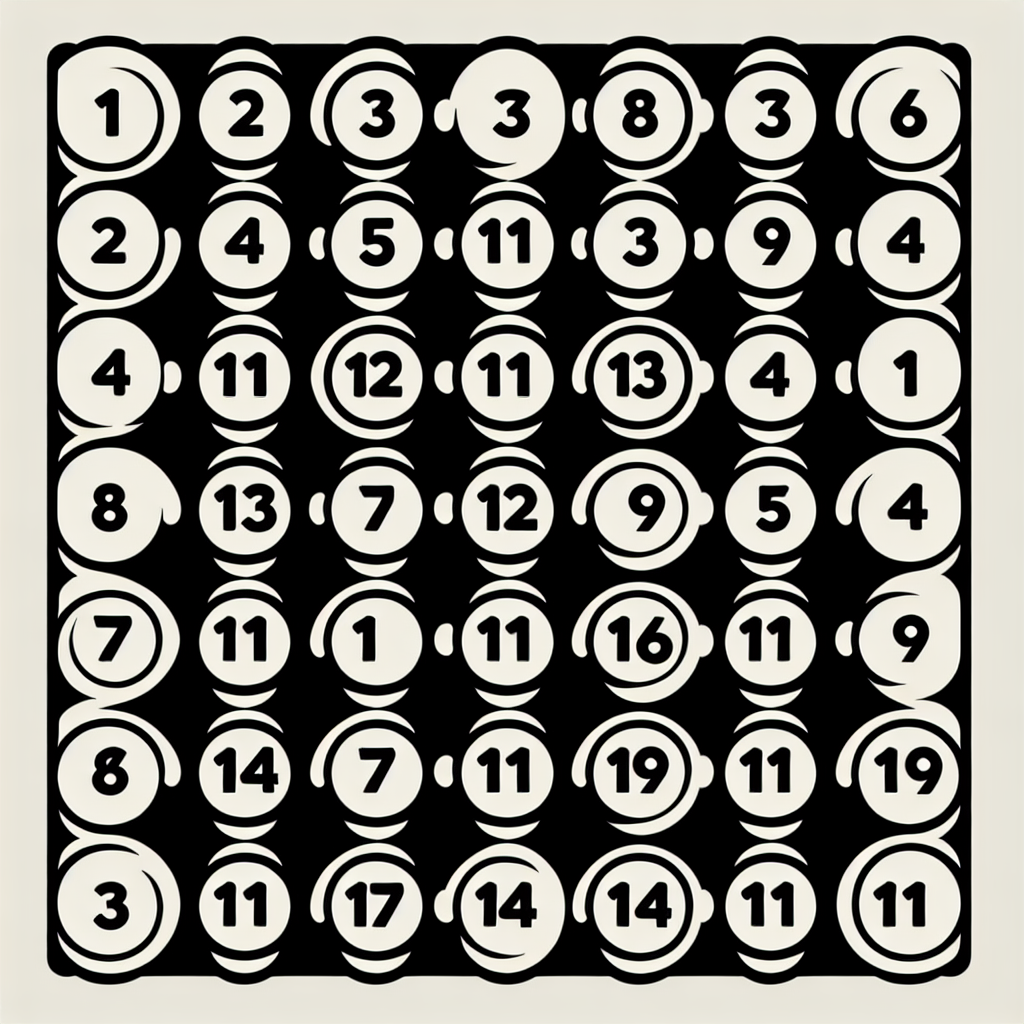} 
        & 2x7 grid of numbered circles, where numbers from 1 to 14 are arranged in seven rows and two columns. \\
        \hline
    \end{tabular}
    \label{tab:pixel_images}
\end{table}

\subsubsection{RQ3: Textual Hint Generation}
We conduct a second ablation experiment, first generating textual hint $T_i$ and then generating diagram $D_i$. This experiment enables us to examine whether LLMs can understand math and anticipate what the hint should be, or need the textual hint to be provided explicitly in the prompt. As shown in Table~\ref{tab:vector_gt_comparison}, the performance drops slightly in this scenario. 

We use the same math problems in Figure~\ref{fig:case_studies} in this analysis. We found that the generated textual hint usually align with the intended explanations, but may contain subtle issues affecting clarity and effectiveness. The textual hint for the multiplication effectively represents $7 \times 2$ as ``$7$ groups of $2$ objects,'' reinforcing the conceptual understanding of multiplication as shown in the ground truth text. However, there exists a minor discrepancy in referring ``circles'' as ``objects,'' could be more precise for consistency with the generated diagram. The textual hint for comparing fractions are technically correct but includes unnecessary details that might interfere with intuitive understanding for students. The hint explicitly describes that $\frac{1}{2}$ means one-half of the circle is filled and $\frac{2}{2}$ means the entire circle is filled, with no explanation on why $\frac{2}{2}$ represents the full circle.

On the other hand, the generated diagrams, though visually similar to the pipeline generated diagrams, may struggle from rendering issues that might impact their usability. In the comparing fractions example, we see that the diagram generated after generating textual hints first struggles from incorrect sizing due to misconfigured \texttt{<viewbox>} attributes. This misconfiguration results in parts of the diagrams appearing cut off, reducing their effectiveness as visual hint. 

\subsubsection{RQ3: On-Task Demonstration}

\begin{figure}[h]
    \centering
    \begin{subfigure}{0.2\textwidth}
        \centering
        \includegraphics[width=\linewidth]{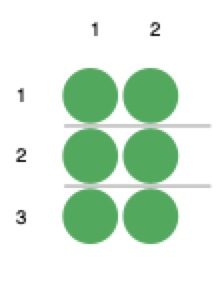}
        \caption{First \\Hint Diagram}
        \label{fig:step_1}
    \end{subfigure}
    \hfill
    \begin{subfigure}{0.2\textwidth}
        \centering
        \includegraphics[width=\linewidth]{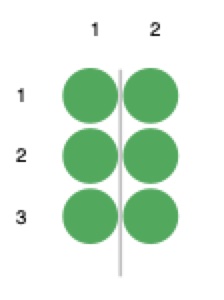}
        \caption{Second \\Hint Diagram}
        \label{fig:step_2}
    \end{subfigure}
    \hfill
    \begin{subfigure}{0.35\textwidth}
        \centering
        \includegraphics[width=\linewidth]{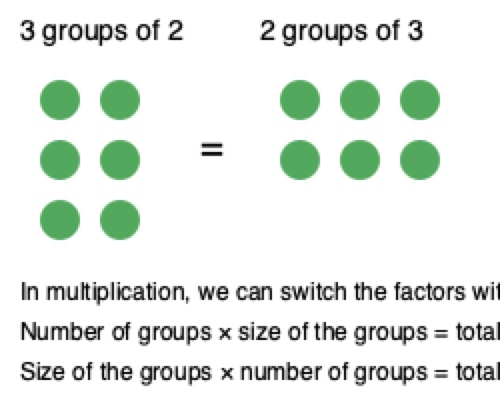}
        \caption{Second \\Hint Diagram (Pipeline)}
        \label{fig:step_2_generated}
    \end{subfigure}
    \caption{A math problem on the topic ``commutative property of multiplication'': (a) shows the first hint diagram, (b) shows the second hint diagram, and (c) shows the second hint diagram generated by the pipeline.}
    \label{fig:diagram_continuity}
\end{figure}

We conduct a third ablation experiment to assess whether an LLM can generate a diagram (\(D_i\)) that aligns with the ground truth when the previous hint step’s diagram (\(D_{i-1}\)) is included in the input prompt. Therefore, we select five problems that have multiple hint steps with diagrams, across three topics: commutative property of multiplication, comparing fractions with the same denominator, and comparing with unit squares.

The generated diagrams show that LLMs can use the previous hint step's diagram to help generate a diagram that closely aligns with the ground truth. Figure~\ref{fig:diagram_continuity} shows an example using the problem: \textit{``Complete the equation. $3 \times 2 = 2 \times ?$''} The first diagram shows green circles in an array structure aligning with the textual feedback \textit{``$3 \times 2$ is equal to $3$ groups of $2$.''} The task is to generate the next diagram corresponding to textual feedback \textit{``$3$ groups of $2$ has the same total as $2$ groups of $3$. Does this always work? Yes! In multiplication, we can switch the factors without changing our total. Number of groups $\times$ size of the groups = total, Size of the groups $\times$ number of groups = total. We call that pattern the commutative property of multiplication .''} We see that the generated diagram for step 2 correctly represents two groups of three. Additionally, it reinforces the equivalence between $3 \times 2$ and $2 \times 3$ by visually grouping circles accordingly. The LLM also annotates the diagram with textual feedback, making it more clear. 

%
These experiments give us valuable insights into the crucial role of context in generating accurate and effective SVG-based diagrams.
In textual hint generation, relying on an LLM to first generate a textual hint and then a diagram is challenging. The model struggles to produce clear textual hints and, consequently, fails to generate well-structured diagrams. This result underscores the importance of textual hints as crucial context for diagram generation.
In on-task demonstration, using the diagram from the previous hint step as an on-task demonstration ensures continuity in the generated diagram, aligning it with the ground truth. This result underscores that the previous hint step serves as crucial context for diagram generation.

\section{Conclusions and Future work}
In this paper, we explored leveraging LLMs to generate hint diagrams for math problems. We found it promising to generate diagrams using Scalable Vector Graphics code through in-context learning and evaluating them with Visual Question Answering (VQA). While our pipeline successfully generated diagrams for certain math topics and diagram types, the LLM often struggles when not enough textual and visual information is present in the input context. 

There are many avenues for future work. 
First, exploring alternative diagram formats could offer greater flexibility and precision in mathematical visualization. Prior research on TikZ-based diagram generation~\cite{belouadi2023automatikz} and Wolfram code-based symbolic representations~\cite{cai2024geogpt4v} suggests that these formats may provide improved rendering for complex mathematical structures. A comparative study of SVG, TikZ, and Wolfram code could highlight the strengths and weaknesses of each format in different educational contexts.
Second, fine-tuning LLMs specifically for diagram generation could significantly enhance output quality. One potential approach is to generate synthetic training data tailored to hint diagrams, incorporating a diverse range of mathematical problem types and ensuring accurate diagram representations. This approach could help mitigate inconsistencies and improve the adaptability of LLMs across different math topics.
Third, integrating human feedback into the diagram generation process could enhance both accuracy and pedagogical effectiveness. Prior research on human-in-the-loop methodologies for math problem generation~\cite{lee2024math} demonstrates the value of educator involvement in refining AI-generated content. Applying similar principles to diagram generation could enable teachers to provide corrective feedback, ensuring that the generated diagrams align with instructional goals and enhance student understanding. 
Fourth, enhancing the robustness of VQA models could further improve mathematical reasoning and diagram interpretation. Expanding training datasets to include a broader range of math-specific visual questions and answers may help VQA models generalize across different problem types, diagram styles, and levels of mathematical complexity.

\newpage
\bibliographystyle{plain}
\bibliography{main}

\end{document}